\documentclass{article}

\usepackage[final]{neurips_2021}
\usepackage[numbers]{natbib}

\usepackage[utf8]{inputenc} % allow utf-8 input
\usepackage[T1]{fontenc}    % use 8-bit T1 fonts
\usepackage{hyperref}       % hyperlinks
\usepackage{url}            % simple URL typesetting
\usepackage{booktabs}       % professional-quality tables
\usepackage{amsfonts}       % blackboard math symbols
\usepackage{nicefrac}       % compact symbols for 1/2, etc.
\usepackage{microtype}      % microtypography
\usepackage{xcolor}         % colors

\usepackage{amsthm}
\theoremstyle{definition}
\newtheorem{definition}{Definition}[section]
\usepackage{multirow}
\usepackage{subcaption}
\usepackage[para]{threeparttable}
\usepackage{graphicx}
\usepackage{float}
\usepackage{amsmath}

\title{Debiasing Credit Scoring using Evolutionary Algorithms}

\author{%
  Nigel Kingsman \\
  University College London, London, UK\\
  Holistic AI, London, UK\\
  \texttt{nigel.kingsman.20@ucl.ac.uk} \\
}

\begin{document}

\maketitle

\begin{abstract}
%   The financial services industry has witnessed the widespread adoption of Artificial Intelligence (AI) in recent years including the deployment of AI for the determination of the creditworthiness of individuals (`credit scoring'). With such AI systems able to deliver outcomes that discriminate against certain groups with protected characteristics, this paper examines the tension between competing training objectives for an AI model where at least one of those objectives is to minimise or decrease the discriminatory bias exhibited by the model.
%   Moreover, this paper presents a methodology for precise model selection when faced with such tension, and the associated difficulties therein.
This paper investigates the application of machine learning when training a credit decision model over real, publicly available data whilst accounting for ``bias objectives". We use the term ``bias objective" to describe the requirement that a trained model displays discriminatory bias against a given groups of individuals that doesn't exceed a prescribed level, where such level can be zero. This research presents an empirical study examining the tension between competing model training objectives which in all cases include one or more bias objectives.\\

This work is motivated by the observation that the parties associated with creditworthiness models have requirements that can not certainly be fully met simultaneously. The research herein seeks to highlight the impracticality of satisfying all parties' objectives, demonstrating the need for ``trade-offs" to be made. The results and conclusions presented by this paper are of particular importance for all stakeholders within the credit scoring industry that rely upon artificial intelligence (AI) models as part of the decision-making process when determining the creditworthiness of individuals. This paper provides an exposition of the difficulty of training AI models that are able to simultaneously satisfy multiple bias objectives whilst maintaining acceptable levels of accuracy. Stakeholders should be aware of this difficulty and should acknowledge that some degree of discriminatory bias, across a number of protected characteristics and formulations of bias, cannot be avoided.
\end{abstract}

% \section{Introduction and impact}
\section{Introduction}

In recent years, the financial services industry has witnessed widespread adoption of artificial intelligence (AI) \citep{jung2019machine}. To date, the implementation of AI has been primarily concerned with capability and accuracy. However, the risks inherent in AI are coming to the fore for research and debate \citep{koshiyama2021towards} noting that AI-driven decision making can deliver outcomes that discriminate against certain groups with protected characteristics. In particular, the promotion of non-discrimination and gender equality, together with other fundamental rights, on the part of AI systems used for credit scoring now forms part of the regulatory agenda \citep{actproposal}.

It is also anticipated that end clients will increasingly scrutinise the discriminatory bias exhibited by AI-driven credit scoring systems, with \citep{kerr2020expectations} noting that there is an expectation by the public that `ethical AI' is possible. Interest in the discriminatory bias shown by AI-driven credit scoring could extend to both the financial and non-financial media together with product comparison sites resulting in a wide range of parties seeking non-discrimination across a range of protected characteristics and formulations of bias. Concurrently, credit scoring institutions will continue to seek to maintain businesses that are financially viable.

In this paper, we report on our findings when exploring multi-objective optimisation problems over a set of publicly available credit datasets. Using the term ``bias objective" to describe the requirement that a trained model displays discriminatory bias against groups of individuals that doesn't exceed a prescribed level, where such level can be zero, we report on the impact on model performance when optimising for bias objectives when using two or more definitions of bias at the same time over a single protected characteristic. We also report on the impact when optimising for bias objectives in the event that more than a single protected characteristic is considered at the same time. We use evolutionary algorithms (EAs) for all our problems, availing of the capability of EAs to operate over non-convex, non-smooth objective functions and to return multiple candidate solutions at once.

% Finally, we also report on a methodology that a model owner might use to select a single solution from the candidate solutions generated by an EA, and the risks inherent with such an approach.

% \paragraph{Impact}

% The results and conclusions presented by this paper of of particular importance for all stakeholders within the credit scoring industry that rely upon AI models as part of the decision-making process when determining the creditworthiness of individuals. This paper provides an exposition of the difficulty of training AI models that are able to simultaneously satisfy multiple bias objectives whilst maintaining acceptable levels of accuracy. Stakeholders should be aware of this difficulty and should acknowledge that some degree of discriminatory bias, across a number of protected characteristics and formulations of bias, cannot be avoided.
This paper comprises the following sections:
\begin{itemize}
    \item \textbf{Background and Related Work:} Certain relevant aspects are reviewed from the literature. We start by providing some of the legislative context concerning fairness more widely in society and review some of the communication on the part of financial regulators highlighting the importance of AI fairness in the domain of credit scoring. We then review the definitions of bias as studied by certain recent papers in the domain of AI fairness more broadly, before reviewing the application of bias objectives in the construction of AI-trained credit scoring models.
    \item \textbf{Experimental Design and Results:} Using evolutionary algorithms, we perform optimisation over problems that simultaneously contain multiple bias objectives. Our experiments reveal that a range of solutions are available for a given multi-objective optimisation problem and provide a good exposition of the tension between the competing objectives of model robustness in terms of accuracy and the reduction of bias across protected characteristics and bias measures.
    \item \textbf{Conclusion and Future Work:} This final section provides the  conclusion of our research and summarises the key findings, before finalising with a brief discussion of future research that could follow this paper.  
\end{itemize}

\section{Background and Related Work}

To provide suitable context for the research presented in this paper, we provide a survey of the related background and literature in three parts: (1) AI fairness legislation, demonstrating the legislative and regulatory desire for AI to be ``fair"; (2) review of bias definitions, exhibiting the breadth of research interest that the conversation around fairness has generated; and (3) application of bias objectives in AI, highlighting the empirical work to date that seeks to mitigate unfair discrimination exhibited by AI models over credit datasets.

\subsection{AI fairness legislation}\label{subsection:fairnessLegislation}

At a global level, the UN's Universal Declaration of Human Rights \citep{assembly1948universal} promotes non-discrimination across a number of attributes. Although not legally binding, the principals and rights of the Declaration have been enshrined in law in a number of countries via national or international legal instruments. Examples of national and international legislation promoting non-discrimination include the United States' Civil Rights Act of 1964 \citep{CRA1964}, the Equality Act 2010 \citep{act2010equality} (which itself consolidated previous legislation) in the UK, and the EU's Equal Pay Directive of 1975 \citep{EP1975}.
% Examples of national and international legislation promoting non-discrimination include the Civil Rights Act of 1964 \citep{CRA1964} and the Americans with Disabilities Act of 1990 \citep{ADA1990} in the United States, the Equality Act 2010 \citep{act2010equality} (which itself consolidated previous legislation) in the UK, the EU's Equal Pay Directive of 1975 \citep{EP1975} and Equal Treatment Directive of 1976 \citep{ET1976}, Article 14 of the Constitution of Japan \citep{Act14} and Articles 14-17 of the Indian Constitution \citep{bakshi1982constitution}.

With regards to the promotion of non-discrimination particularly with respect to algorithms used for credit scoring, a 2016 White House report \citep{executive2016big} called for domains such as credit scoring to design their algorithmic systems so as to build in equal opportunity, whilst in 2020 the UK's Centre for Data Ethics and Innovation (CDEI) highlighted the risk of bias when using algorithms, making particular reference to the risk of bias in credit scoring \citep{cdei2020reviewBias}. Finally, the European Commission set out a proposal in April 2021 for laying down EU-wide rules on AI \citep{actproposal} that enhances and promotes the protection of the rights enshrined in the EU Charter of Fundamental Rights, including rights with respect to non-discrimination. Under the proposal, AI systems used for credit scoring will have to comply with certain mandatory requirements for trustworthy AI and will be subjected to assessment procedures.

Taken together, we can observe an evident desire on the part of lawmakers and financial regulators to ensure that AI is free of bias, and specifically so with regards to the domain of credit scoring.

\subsection{Review of bias definitions} 

During the past decade, there has been a large amount on interest in constructing decision models that are free of bias which has led to researchers constructing a number of bias definitions with accompanying formal mathematical expressions. Moreover, with machine learning (ML) progressively being used more widely and used across multiple domains such as education and healthcare for which outcomes are of massive importance to both individuals and society more widely, there has been a commensurate increasing body of literature that focuses on the application of fairness in ML.

We present a summary of the bias definitions from our review in Table \ref{table:biasDefinitions}. Where applicable and when assuming consideration of the binary case only (that is, we only consider the application to scenarios comprising precisely two outcome possibilities and precisely two prediction possibilities), the table provides an ``equivalent notion of fairness" for each of the bias definitions, availing of the following abbreviations: DI - Disparate Impact, EO - Equal Opportunity, DM - Disparate Mistreatment, and DT - Disparate Treatment.

\begin{table}
  \caption{Summary of bias definitions}
  \label{table:biasDefinitions}
  \centering
  \scalebox{0.9}{
  \begin{tabular}{lll}
  \toprule
Bias definition & \begin{tabular}[c]{@{}l@{}}Literature\\ reference\end{tabular} & \begin{tabular}[c]{@{}l@{}}Equivalent notion of fairness\\ (where applicable)\end{tabular} \\
\midrule
Disparate Impact (``80\% rule") & \citep{feldman2015certifying} & DI\\
$\epsilon$-fairness & \citep{feldman2015certifying} & \\
Equalized Odds & \citep{hardt2016equality} & EO + DM (FPR)\\
Equal Opportunity & \citep{hardt2016equality} & \\
Calibration & \citep{chouldechova2017fair} & DM (FOR) + DM (FDR)\\
Predictive Parity & \citep{chouldechova2017fair} & DM (FDR)\\
Error Rate Balance & \citep{chouldechova2017fair} & DM (FPR) + DM (FNR)\\
Statistical Parity & \citep{chouldechova2017fair} & DI\\
Conditional Statistical Parity & \citep{corbett2017algorithmic} & \\
Predictive Equality & \citep{corbett2017algorithmic} & DM (FPR)\\
Disparate Impact & \citep{zafar2019fairness} & \\
Disparate Treatment & \citep{zafar2019fairness} & \\
Disparate Mistreatment & \citep{zafar2019fairness} & \\
$\epsilon$-general Fair & \citep{oneto2020general} & \\
$\epsilon$-loss General Fair & \citep{oneto2020general} & \\
$\epsilon$-fair & \citep{donini2018empirical} & EO\\
Fairness Through Unawareness & \citep{kusner2017counterfactual} & DT\\
Demographic Parity & \citep{kusner2017counterfactual} & DI\\
Equality of Opportunity & \citep{kusner2017counterfactual} & EO\\
Individual Fairness & \citep{kusner2017counterfactual} & \\
Equal Accuracy & \citep{mitchell2018prediction} & DI\\
Equality of Opportunity & \citep{koshiyama2021towards} & DM (FPR) + EO\\
Equality of Outcome & \citep{koshiyama2021towards} & DI\\
\bottomrule
\end{tabular}
}
\end{table}

It can be seen from Table \ref{table:biasDefinitions} that the majority of the bias definitions from our review can be distilled down to an abridged subset comprising \textit{Disparate Impact}, \textit{Equal Opportunity} and \textit{Disparate Mistreatment}. With $y \in \{-1,1\}$ as the binary class to be predicted ($y=1$ being the ``positive" outcome) and $\hat{y} \in \{-1,1\}$ denoting the prediction of a model, and with $z=0$ denoting the protected group and $z=1$ the unprotected group, we can write this subset of definitions formally as follows:

\begin{definition}[Disparate Impact]\label{defDI} \citep{zafar2019fairness}
A model that predicts $\hat{y}$ satisfies \textit{disparate impact} if the proportion of individuals that are predicted the ``positive" outcome is the same regardless of group membership, formally written as
\begin{equation}
    P(\hat{y}=1|z=0)=P(\hat{y}=1|z=1).
\end{equation}
\end{definition}

\begin{definition}[Equal Opportunity (EO)]\label{defEO} \citep{hardt2016equality}
A model that predicts $\hat{y}$ satisfies \textit{equal opportunity} with respect to $z$ and $y$ if
\begin{equation}
    P(\hat{y}=1|z=0,y=1)=P(\hat{y}=1|z=1,y=1).
\end{equation}
\end{definition}

\begin{definition}[Disparate Mistreatment]\label{defDM} \citep{zafar2019fairness}
A model that predicts $\hat{y}$ does not suffer from \textit{disparate mistreatment} if the model's error rates, with regards to predicting the correct outcome, is the same regardless of group membership. With there being many ways in which error rates can be measured, \citep{zafar2019fairness} provides mathematical definitions for the following specific misclassification measures:
\begin{itemize}
    \item \textbf{Overall misclassification rate (OMR):}
    \begin{equation}
        P(\hat{y}\ne y|z=0)=P(\hat{y}\ne y|z=1),
    \end{equation}
    \item \textbf{False positive rate (FPR):}
    \begin{equation}
        P(\hat{y}\ne y|y=-1,z=0)=P(\hat{y}\ne y|y=-1,z=1),
    \end{equation}
    \item \textbf{False negative rate (FNR):}
    \begin{equation}
        P(\hat{y}\ne y|y=1,z=0)=P(\hat{y}\ne y|y=1,z=1),
    \end{equation}
    \item \textbf{False omission rate (FOR):}
    \begin{equation}
        P(\hat{y}\ne y|\hat{y}=-1,z=0)=P(\hat{y}\ne y|\hat{y}=-1,z=1),
    \end{equation}
    \item \textbf{False discovery rate (FDR):}
    \begin{equation}
        P(\hat{y}\ne y|\hat{y}=1,z=0)=P(\hat{y}\ne y|\hat{y}=1,z=1).
    \end{equation}
\end{itemize}
\end{definition}

\subsection{Application of bias objectives in AI}

The construction of ML-trained models that, in addition to the typical training objective (which usually primarily consists of the minimisation of a loss function), also seeks to meet bias objectives, necessarily requires changes to the typical training pipeline. The same interest that has driven the formulation of a wide range of bias definitions has similarly given rise to a rich research literature concerning the construction of ML-trained models that account for bias objectives. The methodologies for model constructions are broadly split across three categories: \begin{itemize}
    \item \textbf{Pre-processing:} Data is pre-processed prior to being fed into the (unmodified) machine learning algorithm so as to remove the bias being targeted, or otherwise only a subset of features (for example, sensitive information is removed) are passed through to the algorithm for model training.
    \item \textbf{In-processing:} During the model training phase, either constraints or objectives are passed to the (necessarily modified) machine learning algorithm in order to achieve optimisation over the bias objectives in addition to standard training objectives, or an adversary is used during training in order to remove bias.
    \item \textbf{Post-processing:} Outputs from the model generated by the (unmodified) machine learning algorithm are post-processed in order to implement the bias objectives.
\end{itemize}

Table \ref{table:fairnessImplementationsCreditDataSets} highlights the methodologies observed in our review of the literature concerned with building fair AI models over credit datasets. In all cases, other than the two post-processing methodologies listed, in-processing methodologies were used.

\begin{table}
\centering
\begin{threeparttable}
    \caption{Summary of papers: bias objective implementation on credit datasets}
    \scriptsize
    % \begin{center}
    % \centering
    % \scalebox{0.75}{
    \begin{tabular}{llllll}
    \toprule
    \begin{tabular}[c]{@{}l@{}}Literature\\ reference\end{tabular} & ML model & \begin{tabular}[c]{@{}l@{}}Bias definitions\\ (actual or equivalent)\end{tabular} & Bias objective methodology & \begin{tabular}[c]{@{}l@{}}Credit\\ datasets \end{tabular} & \begin{tabular}[c]{@{}l@{}}Sensitive\\features\end{tabular}\\ \midrule
    \citep{kamishima2011fairness} & LR & Disparate Impact & Regularisation approach & Adult & Gender\\
    \citep{hardt2016equality} & N/A & Equalized Odds & Equalised odds post-processing & TransUnion & Race \\
    \citep{hardt2016equality} & N/A & Equal Opportunity & Equal opportunity post-processing & TransUnion & Race \\
    \citep{quadrianto2017recycling} & SVM & Disparate Mistreatment (OMR) & Regularisation approach & Adult & Gender\\
    \citep{zhang2018mitigating} & LR & Equalized Odds & Adversarial Debiasing & Adult & Gender\\ 
    \citep{zafar2019fairness} & LR, SVM & Disparate Impact & Fairness constraint & Adult & Gender, Race\\
    \citep{zafar2019fairness} & LR, SVM & Disparate Impact & Fairness constraint & Bank & Age\\ 
    \citep{donini2018empirical} & SVM & $\epsilon$-fair & Fairness constraint & Adult & Gender\\
    \citep{donini2018empirical} & SVM & $\epsilon$-fair & Fairness constraint & German & Foreign\\ \bottomrule
    \end{tabular}
    % }
    \begin{tablenotes}[para]
    \footnotesize
    \item LR = logistic regression, SVM = support vector machine
    \end{tablenotes}
    % \end{center}
    
    \label{table:fairnessImplementationsCreditDataSets}
\end{threeparttable}
\end{table}

One of the papers cited in Table \ref{table:fairnessImplementationsCreditDataSets}, \citep{zafar2019fairness}, discusses the application of a ``business necessity" clause. The paper's authors observe that the application of bias objectives can yield a trained model that suffers from poor accuracy and propose that a model owner might wish to train a credit scoring model that seeks to optimise over one or more bias objectives whilst being subject to an accuracy constraint in order to ensure that the model achieves a minimum owner-specified performance (the ``business necessity"). They further note that seeking to apply a business necessity clause in the presence of a \textit{disparate mistreatment} bias objective yields a problem comprising a convex-concave objective with convex constraints, a problem that cannot be solved using standard convex solvers, but which might be re-framed as a multi-objective optimisation problem to be tackled using evolutionary algorithms.

This paper generates results using models trained using evolutionary algorithms both to benefit from the flexibility offered by EAs in placing no limitations on the objective functions that might be used and to benefit from EAs' ability to simultaneously generate multiple candidate solutions across a range of model performances (accuracy).

% \section{Exploration of Multi-Objective Optimisation over Credit Datasets}\label{section:multi}
\section{Experimental Design and Results}\label{section:multi}

% In this section, we present both our experimental design and results. 
We investigate the impact on credit model performance due to the simultaneous application of multiple bias objectives whilst training linear models using the Covariance Matrix Adaptation Evolution Strategy (CMA-ES) \citep{igel2007covariance}, an EA considered to be ``state-of-the-art" \citep{cinalli2020hybrid}.

We first train unconstrained logistic regression models over the credit datasets Adult, Bank, German and Mortgage to generate baseline accuracy and bias measures. Using a train/test split of $80/20$ and computing over $20$ runs, we achieve results as per Table \ref{table:unconstrainedTrain}. Note that we use the term ``bias measure" to refer to the quantity $|P(\cdot|\cdot,z=0)-P(\cdot|\cdot,z=1)|$ where the corresponding bias definition would be captured mathematically as $P(\cdot|\cdot,z=0)=P(\cdot|\cdot,z=1)$. Note also that, in our research, the training data results and test data results yielded near-identical conclusions - however, for concision we discuss the training data results only, although we could have equally discussed the test data results.

% \begin{table}
% \caption{Unconstrained results over test data}
% \begin{center}
% \scalebox{0.9}{
% \begin{tabular}{llcccc}
% \toprule
% Dataset & \begin{tabular}[c]{@{}l@{}}Sensitive\\ Attribute\end{tabular} & Mean $\pm$ Std & & & \\ \cmidrule{3-6} 
% & & Accuracy & Disparate & Equal & Disparate \\
% & & & Impact & Opportunity & Mistreatment (OMR)\\ \midrule
% Adult & Gender & 0.8371 $\pm$ 0.0027 &  0.1950 $\pm$ 0.0090 &  0.1758 $\pm$ 0.0307 &  0.1209 $\pm$ 0.0037 \\
% & Race & &  0.1011 $\pm$ 0.0093 &  0.1007 $\pm$ 0.0303 &  0.0545 $\pm$ 0.0099 \\ \midrule
% Bank & Age & 0.9114 $\pm$ 0.0042 &  0.2035 $\pm$ 0.0142 &  0.1493 $\pm$ 0.0350 &  0.1544 $\pm$ 0.0167 \\ \midrule
% German & Age & 0.7678 $\pm$ 0.0294 &  0.1041 $\pm$ 0.0749 &  0.1170 $\pm$ 0.1046 &  0.0634 $\pm$ 0.0521 \\
% & Gender & &  0.0971 $\pm$ 0.0549 &  0.1592 $\pm$ 0.1200 &  0.0518 $\pm$ 0.0470 \\ \midrule
% Mortgage & Gender & 0.9082 $\pm$ 0.0010 &  0.0689 $\pm$ 0.0055 &  0.0292 $\pm$ 0.0042 &  0.0191 $\pm$ 0.0028 \\
% & Race & &  0.1561 $\pm$ 0.0071 &  0.0629 $\pm$ 0.0062 &  0.0046 $\pm$ 0.0025 \\
% \bottomrule
% \end{tabular}
% }
% \end{center}
% \label{table:unconstrainedTest}
% \end{table}

\begin{table}
\caption{Unconstrained results over training data}
\begin{center}
\scalebox{0.9}{
\begin{tabular}{llcccc}
\toprule
Dataset & \begin{tabular}[c]{@{}l@{}}Sensitive\\ Attribute\end{tabular} & Mean $\pm$ Std & & & \\ \cmidrule{3-6} 
& & Accuracy & Disparate & Equal & Disparate \\
& & & Impact & Opportunity & Mistreatment (OMR)\\ \midrule
Adult & Gender & 0.8378 $\pm$ 0.0008 &  0.1952 $\pm$ 0.0020 &  0.1686 $\pm$ 0.0062 &  0.1224 $\pm$ 0.0010\\
& Race & &  0.1025 $\pm$ 0.0031 &  0.1002 $\pm$ 0.0068 &  0.0563 $\pm$ 0.0025\\ \midrule
Bank & Age & 0.9104 $\pm$ 0.0011 &  0.2171 $\pm$ 0.0036 &  0.1773 $\pm$ 0.0074 &  0.1517 $\pm$ 0.0045\\ \midrule
German & Age & 0.7869 $\pm$ 0.0088 &  0.0828 $\pm$ 0.0379 &  0.0526 $\pm$ 0.0462 &  0.0496 $\pm$ 0.0201\\
& Gender & &  0.0982 $\pm$ 0.0298 &  0.0983 $\pm$ 0.0516 &  0.0306 $\pm$ 0.0182\\ \midrule
Mortgage & Gender & 0.9079 $\pm$ 0.0003 &  0.0697 $\pm$ 0.0014 &  0.0289 $\pm$ 0.0009 &  0.0189 $\pm$ 0.0008\\
& Race & &  0.1542 $\pm$ 0.0017 &  0.0616 $\pm$ 0.0013 &  0.0037 $\pm$ 0.0010\\
\bottomrule
\end{tabular}
}
\end{center}
\label{table:unconstrainedTrain}
\end{table}

Table \ref{table:unconstrainedTrain} reveals that, in the unconstrained case, all bias measures (other than \textit{disparate mistreatment} for the Mortgage dataset and sensitive attribute race) yield values greater than $0.01$ and thus we set $0.01$ as a suitable target for our bias measures as we seek to understand the impact upon model accuracy of applying bias objectives.

\subsection{Impact of multiple bias objectives across a single sensitive attribute}\label{subsection:single}

\begin{figure}
    \centering
        \includegraphics[scale=0.65]{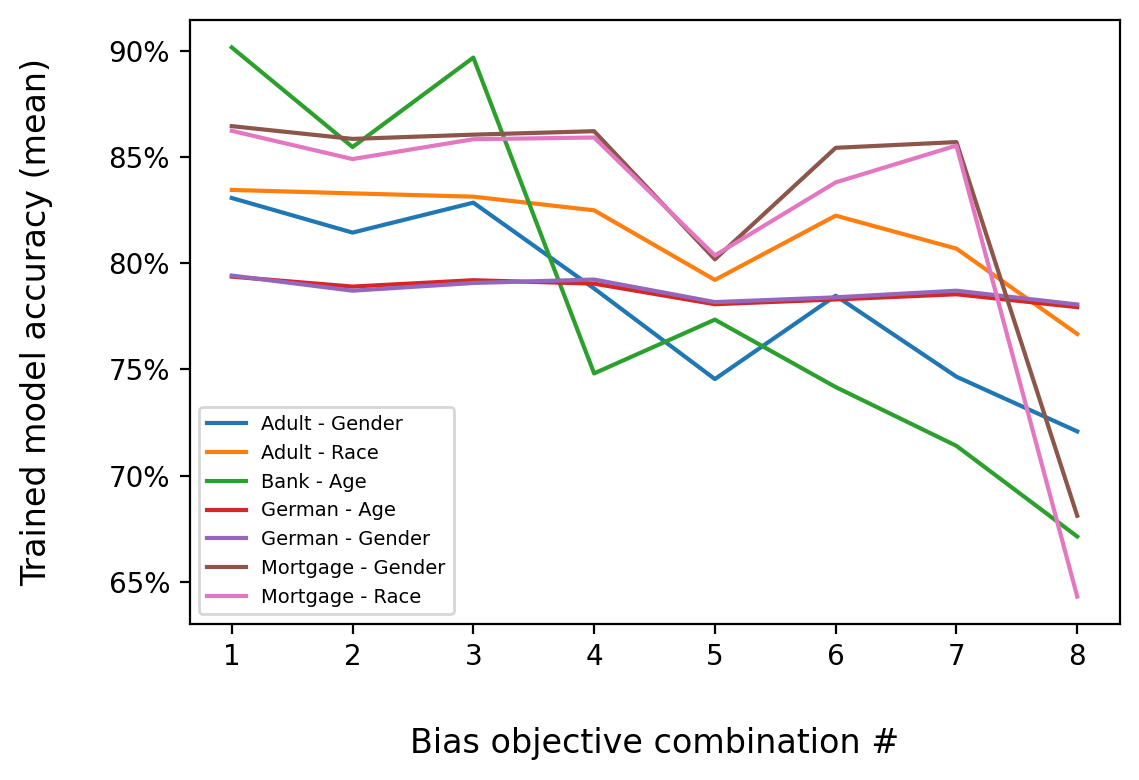}
        \caption{Impact of single attribute bias objectives (training data)}
        \label{fig:problem4test}
\end{figure}

We test the impact of applying a number of bias objectives simultaneously over a single sensitive attribute when training credit scoring models across our four datasets. We use the CMA-ES algorithm to train our models and focus on the impact to model accuracy when enforcing our bias measures to be at most equal to $0.01$. For each of the following bias objective combinations, we record the highest accuracy shown by any of the EA generated candidate solutions that exhibit, over the test data, a measure of at most $0.01$ for the relevant bias objectives:
\begin{enumerate}
    \item No bias objectives met 
    \item Disparate impact (DI) objective met only
    \item Equal opportunity (EO) objective met only
    \item Disparate mistreatment (OMR) (DM) objective met only
    \item Both DI and EO objectives met
    \item Both DI and DM objectives met
    \item Both EO and DM objectives met
    \item All of DI, EO and DM objectives met
\end{enumerate}
Again using a train/test split of $80/20$ over $20$ runs, and running the EA for $10,000$ generations in each case, we get accuracy results as per Figure \ref{fig:problem4test} over training data for each of the dataset/sensitive attribute pairs and for each bias objective combination.

Figure \ref{fig:problem4test} reveals that, of the bias objective combinations that apply two objectives, bias objective combination $5$ (the application of both \textit{disparate impact} and \textit{equal opportunity}) appears to cause the most impact to accuracy, impacting the models trained over the Adult and Mortgage datasets for both of their sensitive attributes. The additional application of the \textit{disparate mistreatment} (OMR) objective (bias objective combination $8$) reduces model accuracy even further.

We note the small impact to model accuracy caused by the imposition of the bias objectives on the models trained over the German dataset, as compared to the larger impacts observed for the models over the other three datasets. Observing that the German dataset is an order of magnitude smaller than any of the other datasets, we briefly hypothesise that the smaller size is in part responsible for this difference in behaviour. To test this, we repeated our experiments over the Bank dataset over the age sensitive attribute, using a subset whose size has been reduced by a factor of $20$ to yield a dataset size of $2,059$ individuals. This reduction in dataset size yields a negative impact to accuracy of approximately $11\%$, as compared to the unconstrained case, once all three bias constraints are applied. This compares to a negative impact of approximately $24\%$ when using the standard, full-size Bank dataset. As such, reducing the size of the Bank dataset, as here, significantly reduces the negative accuracy impact of applying the bias constraints and we thus posit that the comparatively small size of the German dataset might be responsible, at least in part, for the low accuracy impact due to applying the bias objectives, as compared to the other datasets. However, further analysis is certainly needed before any robust conclusions can be made.

\subsection{Impact of bias objectives across multiple sensitive attributes}\label{subsection:multi}

We test the impact of applying a number of bias objectives over two sensitive attributes at once, omitting the Bank dataset for which we only have a single sensitive attribute. Considering the three bias definitions of \textit{disparate impact}, \textit{equal opportunity} and \textit{disparate mistreatment} (OMR), and considering two sensitive attributes, for each of those bias definitions, for each dataset, the objective space naturally takes seven dimensions (including the maximising accuracy objective). To accelerate our experiments, we transform our problem so that the objective space only has four dimensions by collapsing the two bias objectives, one for each sensitive attribute, associated with each bias definition into a single objective by, in each case, minimising the maximum of the bias measure over sensitive attribute $1$ and the bias measure over sensitive attribute $2$. Using the same training pipeline as in Section \ref{subsection:single}, we achieve results as per Table \ref{table:multipleAttrImpact} over the training data.

\begin{table}
\centering
\caption{Impact of multiple attribute bias objectives (training data)}
\scalebox{0.9}{
\begin{tabular}{cccccc}
\toprule
\multicolumn{3}{c}{Dataset} & Adult & German & Mortgage \\ \midrule
\multicolumn{3}{c}{Sensitive Attribute 1} & Gender & Age & Gender \\ \midrule
\multicolumn{3}{c}{Sensitive Attribute 2} & Race & Gender & Race \\ \midrule
DI & EO & DM & \multicolumn{3}{c}{Accuracy (mean $\pm$ std)} \\ \midrule
& & &  0.8319 $\pm$ 0.0019 &  0.7895 $\pm$ 0.0080 &  0.8677 $\pm$ 0.0052 \\
x & & &  0.8138 $\pm$ 0.0018 &  0.7799 $\pm$ 0.0088 &  0.8529 $\pm$ 0.0080 \\
& x & &  0.8205 $\pm$ 0.0111 &  0.7817 $\pm$ 0.0079 &  0.8629 $\pm$ 0.0074 \\
& & x &  0.7851 $\pm$ 0.0040 &  0.7821 $\pm$ 0.0092 &  0.8639 $\pm$ 0.0056 \\
x & x & &  0.6968 $\pm$ 0.0684 &  0.7537 $\pm$ 0.0379 &  0.7449 $\pm$ 0.0589 \\
x & & x &  0.7739 $\pm$ 0.0106 &  0.7707 $\pm$ 0.0119 &  0.8381 $\pm$ 0.0183 \\
& x & x &  0.7112 $\pm$ 0.0269 &  0.7716 $\pm$ 0.0114 &  0.8594 $\pm$ 0.0085 \\
x & x & x &  n/a &  0.7547 $\pm$ 0.0428 &        n/a \\ \bottomrule
\end{tabular}
}
\label{table:multipleAttrImpact}
\end{table}

To understand the impact of meeting our bias objectives across two sensitive attributes simultaneously, we compare the accuracies for each bias combination for the three datasets from Table \ref{table:multipleAttrImpact} against the lowest accuracies obtained over training data across sensitive attributes for the respective dataset and bias combination obtained from our experiment in Section \ref{subsection:single} by subtracting the latter from the former, as given by Table \ref{table:singleVsMultipleAttrImpact}.

\begin{table}
\centering
\caption{Marginal impact of multiple attribute bias objectives (training data)}
\scalebox{0.9}{
\begin{tabular}{cccccc}
\toprule
\multicolumn{3}{c}{Dataset} & Adult & German & Mortgage \\ \midrule
DI & EO & DM & \multicolumn{3}{c}{Accuracy (mean)} \\ \midrule
& & &  0.11\% & -0.42\% & 0.53\% \\
x & & & -0.07\% & -0.72\% & 0.38\% \\
& x & &  -0.81\% & -0.90\% & 0.44\% \\
& & x &  -0.31\% & -0.83\% & 0.46\% \\
x & x & & -4.86\% & -2.70\% & -5.71\% \\
x & & x & -1.09\% & -1.23\% & -0.00\% \\
& x & x & -3.54\% & -1.39\% & 0.39\% \\
x & x & x & n/a &-2.47\% & n/a \\ \bottomrule
\end{tabular}
}
\label{table:singleVsMultipleAttrImpact}
\end{table}

As can be seen in Table \ref{table:singleVsMultipleAttrImpact}, for certain bias objective combinations and datasets, there appears to be a significant negative impact to model accuracy when training the model for bias objectives over two sensitive attributes. In particular, we note that bias objective combination $5$ (both DI and EO objectives are met) yields a negative impact to mean model accuracy of $4.86\%$ and $5.71\%$ for the Adult and Mortgage datasets respectively. Moreover, for those same datasets, when the DM bias objective is also added, we were unable to train a model that adhered to all bias objectives over both sensitive attributes at all, which appears to be a consistent, although more extreme, result. However, we do note that all the model accuracy figures that we are using have some degree of volatility associated with them, and further experiments would need to be conducted to get precision as to the true impact of applying bias objectives over two (or more) sensitive attributes simultaneously. We note that the instances, as per Table \ref{table:singleVsMultipleAttrImpact}, where our experiments imply that the application of bias objectives over two sensitive attributes simultaneously actually yields models that are more accurate than those trained with bias objectives over a single sensitive attribute (for example, over the Mortgage dataset, bias objective combination $4$ yields a mean accuracy that is $0.46\%$ higher) could be similarly as a result of such volatility in the accuracy figures (intuitively, it should not happen that the application of bias objectives over two sensitive attributes yield greater accuracy than over a single sensitive attribute, as the former should just be a special case of the latter).

Notwithstanding the reservations expressed above, our empirical results in Table \ref{table:singleVsMultipleAttrImpact} do suggest some negative impact from the application of bias objectives over two sensitive attributes, with the differences in the table skewed towards the negative.

\subsection{Trade-offs between model objectives}\label{section:tradeOffs}

\begin{figure}
    \centering
        \includegraphics[scale=1.0]{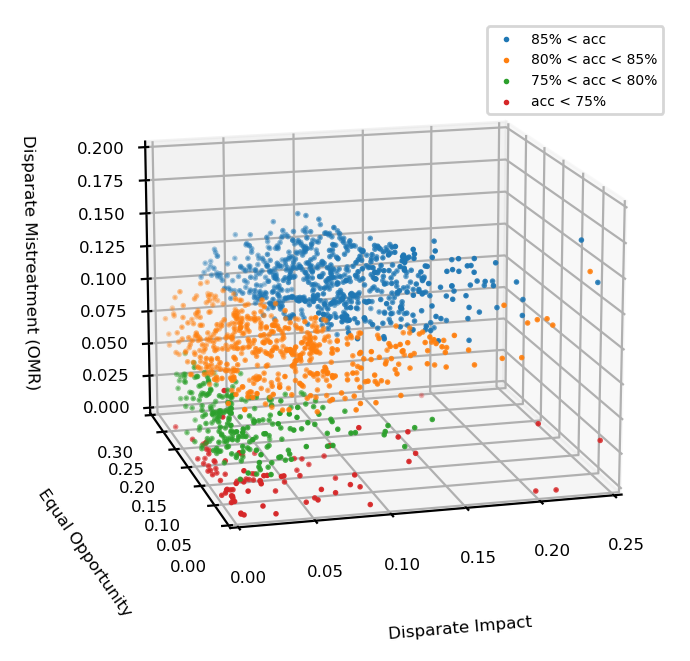} % first Figure itself
        \caption{Best performing run on Bank dataset (training data), bucketed by accuracy}
        \label{fig:problem4_bank_bucketAcc}
\end{figure}

The results in Sections \ref{subsection:single} and \ref{subsection:multi} have highlighted the trade off between model accuracy and bias objectives, with model accuracy suffering as bias objectives were added. We consider here the more general case where any of the model objectives might need to be relaxed as other objectives are imposed or strengthened. Revisiting our results from Section \ref{subsection:single} over the full-sized Bank dataset, we focus upon the CMA-ES solution set that achieved the largest hypervolume in objective space over the $20$ runs. Bucketing the objective space solutions by accuracy yields the 3D plot of Figure \ref{fig:problem4_bank_bucketAcc}.

Figure \ref{fig:problem4_bank_bucketAcc} shows that, as model accuracy increases, the \textit{disparate impact} and \textit{disparate mistreatment} (OMR) measures are forced to take higher values, providing more granularity than Section \ref{subsection:single}'s results. The clusters for the higher accuracy buckets exhibit some visible curvature on the DI-EO plane (i.e. for fixed values of DM (OMR)) suggesting that, for this dataset and for a given accuracy and DM (OMR) measure value, some success on the \textit{disparate impact} measure needs to be sacrificed for success on the \textit{equal opportunity} measure.

Similarly, when we bucket by \textit{disparate mistreatment} (OMR), clear relationships can be observed with the error rate increasing (accuracy decreasing) as \textit{disparate mistreatment} is progressively decreased, as would be expected, whilst the \textit{disparate impact} also simultaneously decreases. That second relationship is unsurprising given the results shown in Figure \ref{fig:problem4test}, where the addition of the \textit{disparate impact} objective on top of the \textit{disparate mistreatment} objective caused only a small marginal reduction in accuracy.

\begin{figure}
    \centering
        \includegraphics[scale=0.7]{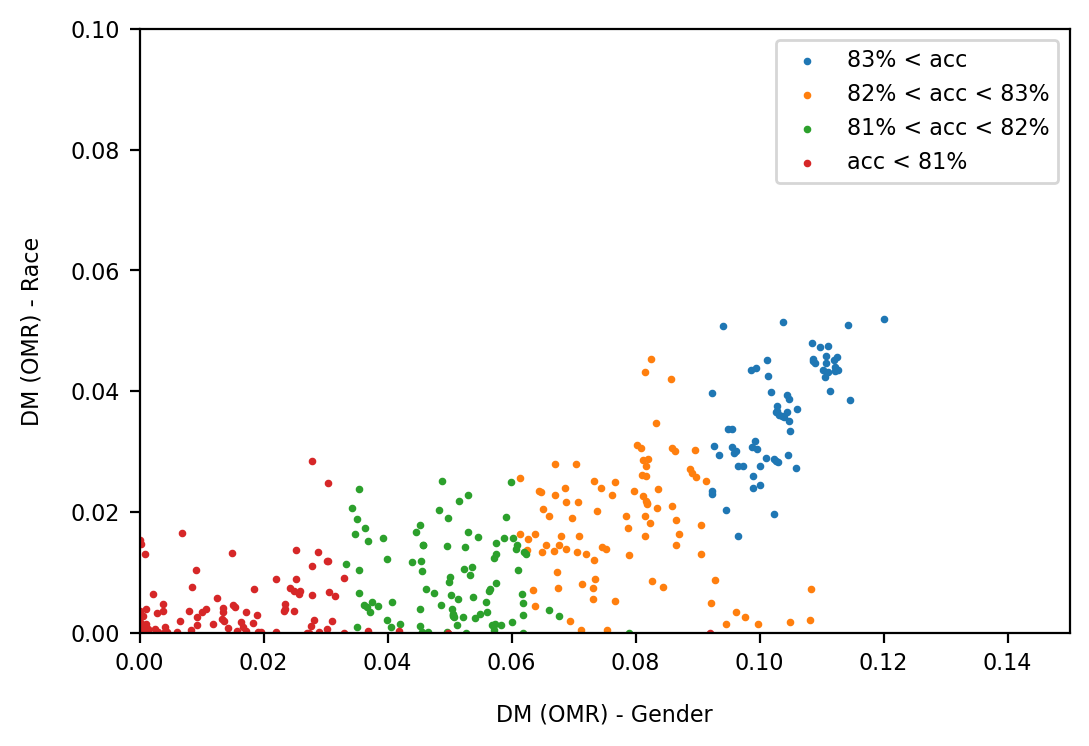} % first Figure itself
        \caption{Best performing run on Adult dataset (training data), bucketed by accuracy}
        \label{fig:problem6_adult_bucketAcc}
\end{figure}

To consider the tension between bias objectives across more than one sensitive attribute, we trained models over the Adult dataset that, in addition to seeking to maximise accuracy, sought to minimise both \textit{disparate mistreatment} (OMR) over gender and \textit{disparate mistreatment} (OMR) over race. This experiment yielded less clear relationships. By way of example, Figure \ref{fig:problem6_adult_bucketAcc} presents the results achieved by the solution set that achieved the largest hypervolume out of 20 runs over the training data when bucketed by accuracy.

Inspection of Figure \ref{fig:problem6_adult_bucketAcc} suggests that there is a small amount of trade-off between the two bias objectives for a fixed accuracy, with the boundaries between each accuracy cluster indicating that the \textit{disparate mistreatment} (OMR) measure over gender needs to be subjected to a small increase in value as the measure over race is decreased in order to maintain a constant accuracy.

% \section{Methodology for Precise Model Selection}\label{section:precise}
\section{Conclusion and Future Work}

We have presented an investigation that considered the impact of simultaneously training a credit model, using an evolutionary algorithm, over a number of bias objectives on model performance (in terms of accuracy). The first part of this investigation showed that progressively adding bias objectives over a single sensitive attribute when training for the model yields a progressive decrease in model performance, although we did observe that such decrease did vary between the datasets studied, with one dataset in particular exhibiting very little performance sensitivity to the addition of bias objectives. The second part of this investigation considered whether the progressive addition of bias objectives over more than a single sensitive attribute yields further model performance decreases. We saw some empirical evidence to suggest that that is indeed the case, although further investigations are required to yield more conclusive results. The investigation concluded with a study of the trade-offs between bias objectives. Only a small amount of such trade-off was observed. This result was surprising as we had made the incorrect assumption that the clear trade-off between model accuracy and bias objectives would be matched by a clear trade-off between the bias objectives themselves.

This research provides an understanding of the trade-offs between objectives that must be considered by credit model owners when seeking to solve multi-objective optimisation problems where such problems contain bias objectives. Furthermore, this research shows that results can be obtained over such problems using evolutionary algorithms for model training, a technique that is little used in the existing literature for applying bias objectives. This research communicates the difficulty, for credit industry practitioners, of maintaining credit models that are robust with regards to accuracy and show little to no bias with respect to individuals' sensitive attributes. The findings presented herein should be of interest to all stakeholders in a credit business including lawmakers, regulators and end clients, as there does need to be a wider understanding of what can and cannot be achieved presently with regards to eliminating discrimination in credit provision.

Finally, we discuss directions for future research. With the EA implementations used in our investigation proving resource heavy from a computational perspective, we restricted ourselves to only investigating the application of bias objectives for up to two sensitive attributes simultaneously. Two of the datasets featured in this paper, German and Mortgage, could potentially allow experimentation over more than two sensitive attributes simultaneously. An interesting line of investigation would be to research whether there was an outsized impact to model accuracy when training over bias objectives over three or more sensitive attributes, for which an investigation of alternative EA implementations (or, indeed, the construction of a new implementation) might be required.

\section*{Acknowledgements}

I would like to thank Philip Treleaven and Adriano Koshiyama for the advice and feedback they have kindly provided on the work that informed the material herein. Moreover, I would like to thank Emre Kazim for his kind advice particular to this paper.

\bibliographystyle{plain}
% \bibliography{references}

\appendix

\section*{Supplementary Material}

\section{Datasets}\label{section:datasets}
This paper refers to four datasets, as detailed below. In all cases, these datasets comprise features that are deemed ``sensitive" and are subject to anti-discrimination laws in one or more jurisdictions. Each dataset also contains an outcome for each individual. The first three of these datasets, `Adult', `Bank' and `German' were sourced from the UCI Machine Learning Repository (\url{http://archive.ics.uci.edu/ml}) \citep{Dua:2019}, whilst the final dataset, `Mortgage' was sourced from \url{https://github.com/askoshiyama/audit\_mortgage} \citep{koshiyama2020Mortgage}.

\subsection*{Adult dataset}
This dataset contains details of $45,222$ individuals comprising $14$ features covering a range of demographic characteristics including education, occupation and personal relationship details. For our studies, we use gender (labelled as `sex' in the raw data) and race as our sensitive attributes. The outcome for the dataset is whether an individual has an income that exceeds $\$50,000$ per annum (the positive outcome, labelled $y=1$) or less than or equal to $\$50,000$ per annum (the negative outcome, labelled $y=-1$).

The dataset has the following distribution of sensitive features and class labels (positive or negative outcome):

\begin{table}[H]
\centering
\caption{Adult dataset - distribution of outcomes by gender}
\begin{tabular}{lccc}
\toprule
Gender & Low income ($y=-1$) & High income ($y=1$) & Total \\ \midrule
Males (non-protected) & 20,988 (68.75\%) & 9,539 (31.25\%) & 30,527 (100.00\%) \\ 
Females (protected) & 13,026 (88.64\%) & 1,669 (11.36\%) & 14,695 (100.00\%) \\ \midrule
Total & 34,014 (75.22\%) & 11,208 (24.78\%) & 45,222 (100.00\%) \\ \bottomrule
\end{tabular}
\label{tab:statsAdultGender}
\end{table}

\begin{table}[H]
\centering
\caption{Adult dataset - distribution of outcomes by race (white/non-white)}
\begin{tabular}{lccc}
\toprule
Race & Low income ($y=-1$) & High income ($y=1$) & Total \\ \midrule
White (non-protected) & 28,696 (73.76\%) & 10,207 (26.24\%) & 38,903 (100.00\%) \\ 
Non-white (protected) & 5,318 (84.16\%) & 1,001 (15.84\%) & 6,319 (100.00\%) \\ \midrule
Total & 34,014 (75.22\%) & 11,208 (24.78\%) & 45,222 (100.00\%)\\ \bottomrule
\end{tabular}

\label{tab:statsAdultRace}
\end{table}

As can be seen from Tables \ref{tab:statsAdultGender} and \ref{tab:statsAdultRace}, there is a discrepancy in actual outcomes between non-protected and protected groups when gender and race are considered. In both cases, the non-protected groups enjoy a higher proportion of positive outcomes than the protected groups.

\subsection*{Bank dataset}
The Bank dataset is a marketing dataset comprising details of $41,188$ individuals each with $20$ features including personal and financial attributes. We use age as the sensitive attribute, with those persons between the ages of $25$ and $60$ (inclusive) deemed to be in the non-protected group, and those persons outside that age range in the protected group. The positive outcome ($y=1$) for this dataset is the successful subscription of an individual to a bank term deposit.

The dataset has the following distribution of the sensitive feature and class labels (positive or negative outcome):

\begin{table}[H]
\centering
\caption{Bank dataset - distribution of outcomes by age}
\begin{tabular}{lccc}
\toprule
Age & No subscription ($y=-1$) & Subscription ($y=1$) & Total \\ \midrule
\begin{tabular}[c]{@{}l@{}}$25\le \text{age} \le 60$\\ (non-protected group)\end{tabular} & 1,308 (66.13\%) & 670 (33.87\%) & 1,978 (100.00\%) \\ 
\begin{tabular}[c]{@{}l@{}}$\text{age}<25\text{ or age}>60$ \\ (protected group)\end{tabular} & 35,240  (89.88\%) & 3,970 (10.12\%) & 39,210 (100.00\%) \\ \midrule
Total & 36,548 (88.73\%) & 4,640 (11.27\%) & 41,188 (100.00\%)\\ \bottomrule
\end{tabular}
\label{tab:statsBankAge}
\end{table}

Again a discrepancy in actual outcomes between the non-protected and protected groups can be observed with respect to the sensitive attribute, with Table \ref{tab:statsBankAge} showing that a greater proportion ($33.87\%$) of the non-protected group receives a positive outcome than the protected group ($10.12\%$).

\subsection*{German dataset}
We use a specific version of this dataset (\url{https://doi.org/10.7910/DVN/Q8MAW8} \citep{DVN/Q8MAW8_2016}) which has undergone pre-processing to convert a number of categorical attributes to numerical attributes. The dataset thus contains details of $1,000$ individuals with $24$ features including those pertaining to financial situation and history together with some personal data. We use age and gender as our sensitive attributes, selected so as to allow comparison to the other datasets presented here. As with the Bank dataset, we convert age into a binary attribute by deeming those persons between the ages of $25$ and $60$ (inclusive) to be in the non-protected group and those persons outside that range to be in the protected group. The positive outcome ($y=1$) for this dataset is the individual being determined as a good credit risk, with the negative outcome a bad credit risk.

The dataset has the following distribution of sensitive features and class labels (positive or negative outcome):

\begin{table}[H]
\centering
\caption{German dataset - distribution of outcomes by age}
\begin{tabular}{lccc}
\toprule
Age & Bad credit risk ($y=-1$) & Good credit risk ($y=1$) & Total \\ \midrule
\begin{tabular}[c]{@{}l@{}}$25\le \text{age} \le 60$\\ (non-protected group)\end{tabular} & 123 (63.40\%) & 71 (36.60\%) & 194 (100.00\%) \\ 
\begin{tabular}[c]{@{}l@{}}$\text{age}<25\text{ or age}>60$ \\ (protected group)\end{tabular} & 577  (71.59\%) & 229 (28.41\%) & 806 (100.00\%) \\ \midrule
Total & 700 (70.00\%) & 300 (30.00\%) & 1,000 (100.00\%)\\ \bottomrule
\end{tabular}
\label{tab:statsGermanAge}
\end{table}

\begin{table}[H]
\centering
\caption{German dataset - distribution of outcomes by gender}
\begin{tabular}{lccc}
\toprule
Gender & Bad credit risk ($y=-1$) & Good credit risk ($y=1$) & Total \\ \midrule
Males (non-protected) & 499 (73.32\%) & 191 (27.68\%) & 690 (100.00\%) \\ 
Females (protected) & 201 (64.84\%) & 109 (35.16\%) & 310 (100.00\%) \\ \midrule
Total & 700 (70.00\%) & 300 (30.00\%) & 1,000 (100.00\%)\\ \bottomrule
\end{tabular}
\label{tab:statsGermanGender}
\end{table}

% \begin{table}[H]
% \centering
% \caption{German dataset - distribution of outcomes by foreign worker status}
% \begin{tabular}{lccc}
% \toprule
% Foreign worker status & Bad credit risk ($y=-1$) & Good credit risk ($y=1$) & Total \\ \midrule
% No (non-protected) & 33 (89.19\%) & 4 (10.81\%) & 37 (100.00\%) \\ 
% Yes (protected) & 667 (69.26\%) & 296 (30.74\%) & 963 (100.00\%) \\ \midrule
% Total & 700 (70.00\%) & 300 (30.00\%) & 1,000 (100.00\%)\\ \bottomrule
% \end{tabular}
% \label{tab:statsGermanForeign}
% \end{table}

Tables \ref{tab:statsGermanAge} and \ref{tab:statsGermanGender} both exhibit a discrepancy in actual outcomes between the non-protected and protected groups, with a greater proportion of the non-protected group receiving a positive outcome when age is considered, and a greater proportion of the protected group receiving a positive outcome when gender is considered (showing a bias against the non-protected group contrary to what we've seen thus far in this section).

\subsection*{Mortgage dataset}

The outcome-balanced Mortgage dataset, comprising data from 2011 collected under the Home Mortgage Disclosure Act (HMDA, \citep{HDMA1975}), contains $200,000$ individuals with $29$ features containing details of the mortgages being requested (for example, amount, mortgage purpose, security over the mortgage) and personal information. We use gender and race as our sensitive attributes to facilitate comparison against the other datasets. The outcome for the dataset is whether an individual is extended a mortgage (the positive outcome, labelled $y=1$) or not (the negative outcome, labelled $y=-1$).

The dataset has the following distribution of sensitive features and class labels (positive or negative outcome):

\begin{table}[H]
\centering
\caption{Mortgage dataset - distribution of outcomes by gender}
\begin{tabular}{lccc}
\toprule
Gender & Withheld ($y=-1$) & Granted ($y=1$) & Total \\ \midrule
Males (non-protected) & 67,838 (48.05\%) & 73,334 (51.95\%) & 141,172 (100.00\%) \\ 
Females (protected) & 32,162 (54.67\%) & 26,666 (45.33\%) & 58,828 (100.00\%) \\ \midrule
Total & 100,000 (50.00\%) & 100,000 (50.00\%) & 200,000 (100.00\%)\\ \bottomrule
\end{tabular}
\label{tab:statsMortgageGender}
\end{table}

\begin{table}[H]
\centering
\caption{Mortgage dataset - distribution of outcomes by race (white/non-white)}
\begin{tabular}{lccc}
\toprule
Race & Withheld ($y=-1$) & Granted ($y=1$) & Total \\ \midrule
White (non-protected) & 82,827 (48.34\%) & 88,517 (51.66\%) & 171,344 (100.00\%) \\ 
Non-white (protected) & 17,173 (59.93\%) & 11,483 (40.07\%) & 28,656 (100.00\%) \\ \midrule
Total & 100,000 (50.00\%) & 100,000 (50.00\%) & 200,000 (100.00\%)\\ \bottomrule
\end{tabular}
\label{tab:statsMortgageRace}
\end{table}

As can be seen in Tables \ref{tab:statsMortgageGender} and \ref{tab:statsMortgageRace}, for both of our sensitive attributes, a greater proportion of the non-protected group receive positive actual outcomes than the protected group.

\end{document}